\documentclass[11pt]{article}
\usepackage{acl}
\usepackage{times}
\usepackage{latexsym}
\usepackage[T1]{fontenc}
\usepackage[utf8]{inputenc}
\usepackage{microtype}
\usepackage{inconsolata}
\usepackage{graphicx}
\usepackage{booktabs}
\usepackage{amsmath}
\usepackage{array}

\title{Push Your Agent: Measuring and Enforcing Quantitative Goal Persistence in Long-Horizon LLM Agents}

\author{
Yuandao Cai\textsuperscript{1} \quad
Yuzhang Zhu\textsuperscript{1} \quad
Liyou Gao\textsuperscript{1} \quad
Wensheng Tang\textsuperscript{1} \quad
Shengchao Qin\textsuperscript{2} \\
\textsuperscript{1}Independent Researcher \quad
\textsuperscript{2}Xidian University
}

\newcommand{\pushbench}{PushBench}
\newcommand{\qgp}{Quantitative Goal Persistence}
\newcommand{\valid}{\texttt{valid\_count}}
\newcommand{\target}{\texttt{target\_count}}
\newcommand{\reported}{\texttt{reported\_count}}
\newcommand{\stateful}{\textsc{StateQGP}}
\newcommand{\workunit}{\textsc{UnitQGP}}
\newcommand{\reposcan}{QGP-RepoScan}
\newcommand{\dataops}{QGP-DataOps-lite}
\newcommand{\swelite}{QGP-SWE-lite}
\newcommand{\letta}{\textsc{Letta/MemGPT}}
\definecolor{methodred}{RGB}{180,35,35}
\newcommand{\methodhi}[1]{\textcolor{methodred}{\textbf{#1}}}

\AtBeginEnvironment{table}{\ifdefined\nolinenumbers\nolinenumbers\fi}
\AtEndEnvironment{table}{\ifdefined\linenumbers\linenumbers\fi}
\AtBeginEnvironment{table*}{\ifdefined\nolinenumbers\nolinenumbers\fi}
\AtEndEnvironment{table*}{\ifdefined\linenumbers\linenumbers\fi}
\AtBeginEnvironment{figure}{\ifdefined\nolinenumbers\nolinenumbers\fi}
\AtEndEnvironment{figure}{\ifdefined\linenumbers\linenumbers\fi}
\AtBeginEnvironment{figure*}{\ifdefined\nolinenumbers\nolinenumbers\fi}
\AtEndEnvironment{figure*}{\ifdefined\linenumbers\linenumbers\fi}

\begin{document}
\maketitle

\begin{abstract}
Long-horizon language agents can make many plausible local tool calls yet fail to persist until a
requested count is actually complete. We study this gap as \qgp{} (QGP): whether an agent keeps
working until an external verifier confirms enough distinct valid items. \pushbench{} turns this into
a benchmark for repository-artifact collection and verifier-backed work units, so repeated work,
duplicate submissions, false completion, and progress drift are measured directly rather than
hidden behind a final success flag. 
In matched controller comparisons, a state-tracking retrieval
controller reaches 69--78\% success while eliminating duplicate submissions, and a backlog-tracking
work-unit controller reaches 25--50\% success in settings where standard and completion-gated
controllers complete no task instances. Black-box frontier-agent evaluations with Claude Code
(Sonnet 4.6) and Codex CLI (\texttt{gpt-5.4}) solve many
50-artifact tasks but drop to 3/9 successes per condition at 100 artifacts. 
The results show that quantitative goals stress a different reliability requirement from
local task competence: agents must maintain verified progress and stop only when the requested
work is complete.
\end{abstract}

\section{Introduction}
Tool-using language agents are now routinely asked to search, collect evidence, edit code, operate
web interfaces, and coordinate many steps before producing an answer. Many such tasks are not
complete when the agent has found \emph{some} useful items or made \emph{one} plausible edit. They
are complete only when an explicit, auditable condition has been met: a backlog cleared, required
records processed, failing checks passed, or enough repository artifacts collected. In these
settings, local competence is not enough. The reliability problem is whether an agent's progress
toward an explicit goal is maintained until the completion condition is satisfied.

Recent agent benchmarks cover increasingly realistic reasoning-and-acting frameworks, web
environments, and repository tasks
\citep{yao2023react,schick2023toolformer,yao2022webshop,zhou2024webarena,jimenez2024swebench}.
They capture task success across interactive settings
\citep{liu2024agentbench,xie2024osworld,ma2024agentboard,yao2024taubench}. However,
final success rates alone often hide how progress was lost. A failed trajectory may reflect poor
local decisions, but it may also reflect premature stopping, repeated work, duplicate submissions,
or a final message that overstates how much verified work has actually been completed. This is
especially problematic when an agent appears helpful for many turns and still stops short of the
requested count.

We study this behavior through \qgp{} (QGP), which asks whether an agent continues until an
external verifier confirms at least $N$ distinct valid work units. This framing makes duplicate
submissions, false completion, premature stopping, and reported-count error measured failure modes
rather than incidental artifacts of failed trajectories.
QGP therefore differs from final success, recall@$N$, or ordinary pass/fail scoring. It also differs
from progress-oriented evaluations that assign partial credit to trajectories
\citep{ma2024agentboard}: QGP makes the stopping condition itself externally checkable, so repeated
submissions, premature stopping, and unsupported completion claims are part of the task outcome.

\begin{figure}[t]
\centering
\includegraphics[width=\columnwidth]{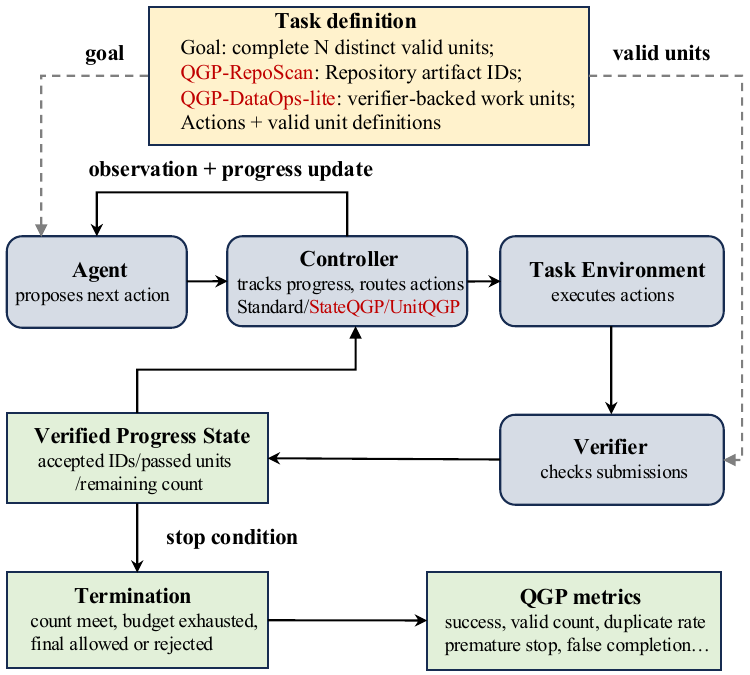}
\caption{\pushbench{} workflow: agents act through a controller, task environment, and verifier until
the count goal is met or the budget is exhausted.}
\label{fig:pushbench-overview}
\end{figure}

\pushbench{} is a benchmark and evaluation framework for this kind of verifier-audited
progress. It contains two controlled task families. Repository-artifact collection (\reposcan{})
isolates counting and retrieval over real repository artifacts. Verifier-backed work units
(\dataops{}) extend the setting to small edits, checks, and structured updates that count only
after a deterministic verifier accepts them. These controlled, verifier-checkable, low-to-medium
difficulty tasks reduce impossible-goal explanations and more directly measure quantitative
persistence.
Note that we also include coding-style and black-box frontier-agent evaluations to test whether
the same progress checks remain useful in more realistic settings, where persistence is intertwined
with coding skill, repository navigation, hidden prompting, tool routing, and product-specific
loops.
Figure~\ref{fig:pushbench-overview} gives the workflow: task definitions provide the
goal and valid-unit criteria, while the controller and verifier maintain progress through repeated
actions. 

We further evaluate controller-level interventions tailored to the two \pushbench{} task families.
For repository-artifact retrieval, a state-tracking retrieval controller (\stateful{}) keeps
externally visible state over submitted identifiers, repeated search pages, and completion
eligibility. For verifier-backed work units, a backlog-tracking controller (\workunit{}) keeps
verifier-visible state over pending, attempted, and passed units. These controllers do not alter the
verifier or the set of valid items; they change how verified progress is exposed, tracked, and
enforced so that an execution cannot repeatedly spend budget on completed or no-progress work.

Across these settings, progress-state mechanisms reduce repeated work and premature stopping.
 In
repository-artifact collection, state-tracking control reaches 69--78\% success under matched
model and backend settings while eliminating duplicate submissions. In verifier-backed work-unit
tasks, standard and completion-gated controllers complete no task instances in this evaluation,
while backlog-tracking control reaches 25--50\% success. Memory agents and frontier systems can
help, but they do not make verified progress automatic: black-box frontier-agent evaluations solve
many 50-artifact tasks but fall sharply at 100 artifacts. Overall, QGP failures remain visible even
when stronger models, memory mechanisms, and frontier-agent harnesses improve absolute performance.
Our contributions are:
\begin{itemize}
    \item We formalize QGP as an evaluation target for auditable progress toward
    explicit count goals.
    \item We design \pushbench{} with repository-artifact and verifier-backed
    work-unit task families.
    \item We evaluate controller-level interventions under matched backends against passive,
    verifier-gated, and memory-agent baselines, with coding-style and black-box frontier-agent
    evaluations testing how the pattern transfers beyond the controlled settings.
\end{itemize}

\section{Related Work}
\paragraph{Agent and software benchmarks.}
Recent benchmarks evaluate agents in interactive web, desktop, data, and tool-using environments
\citep{liu2024agentbench,yao2022webshop,zhou2024webarena,xie2024osworld,yoran2024assistantbench,
yao2024taubench,ma2026dataagentbench,huang2024mlagentbench}; related text-environment and web
navigation tasks include ALFWorld, ScienceWorld, and Mind2Web
\citep{shridhar2021alfworld,wang2022scienceworld,deng2023mind2web}. Software benchmarks similarly
cover code generation, repository context, and issue repair, from HumanEval, MBPP, CodeXGLUE, and
APPS to InterCode, RepoBench, SWE-bench, and SWE-agent
\citep{chen2021codex,austin2021mbpp,lu2021codexglue,hendrycks2021apps,yang2023intercode,
liu2024repobench,jimenez2024swebench,yang2024sweagent}. These benchmarks establish task realism
and executable success; \pushbench{} instead isolates quantitative persistence and externally
auditable stopping conditions.

\paragraph{Tool use and agent control.}
Work on tool use and control improves how language models select tools, interleave reasoning and
action, and recover from intermediate mistakes. ReAct, Toolformer, ToolLLM, API-Bank, Gorilla,
StableToolBench, ToolHop, and tool-use alignment benchmarks study tool invocation and API use
\citep{schick2023toolformer,qin2024toolllm,li2023apibank,patil2024gorilla,
guo2024stabletoolbench,ye2025toolhop,chen2024toolalignment,yao2023react}; middleware,
model-based planning, reflection, and self-feedback work study execution infrastructure and
trajectory repair \citep{gu2024middleware,gu2025worldmodelweb,shinn2023reflexion,madaan2023selfrefine}.
These methods can improve local action choice and recovery, but they typically leave the
quantitative stopping invariant implicit. \pushbench{} studies a different layer: the controller
contract between policy actions and verifier-visible progress state.

\paragraph{Progress, partial credit, and trajectory evaluation.}
AgentBoard proposes fine-grained progress rates for multi-turn agents \citep{ma2024agentboard},
and interactive benchmarks often report partial completion or trajectory-derived signals.
QGP differs in making the stopping condition itself the object of evaluation: false completion,
premature stopping, duplicate work, and reported-count error are failure modes that show whether
externally auditable progress has been maintained. This makes verified progress tracking part of the task
contract rather than a post-hoc trajectory summary.

\paragraph{Memory and orchestration.}
Memory-augmented and long-running agents manage context, user preferences, and evolving state across
interactions, including explicit memory systems, persistent simulated or embodied agents, and
general autonomous-agent architectures
\citep{packer2023memgpt,park2023generativeagents,wang2024voyager,wang2024agentssurvey}. These
systems are natural comparisons because QGP failures can look like memory failures. Our distinction
is that QGP state is not arbitrary recalled context: it is externally checkable progress over
submitted units, duplicates, remaining target count, and termination eligibility.

\section{Quantitative Goal Persistence}
\label{sec:qgp}
We define a QGP task as a tuple $(\mathcal{E}, T, V, N, B)$, where $\mathcal{E}$ is an interactive
environment, $T$ is the task objective, $V(x) \in \{0,1\}$ is an external verifier for a submitted
work unit $x$, $N$ is the target count, and $B$ is the interaction budget. A work unit can be any
verifier-visible item whose identity and validity can be checked, such as a repository backlog
item, issue, data record, file chunk, code location, test case, build target, or queue item. At step $t$,
the agent may inspect the environment, attempt a work unit, submit an identifier or artifact, ask
the user, or produce a final answer. Let $C_t$ be the multiset of submitted or attempted work-unit
identifiers and let $D_t$ be the set of distinct identifiers in $C_t$. The verified progress at
step $t$ is
\begin{equation}
    \valid_t = \left|\{x \in D_t : V(x)=1\}\right|.
\end{equation}
The run is complete if and only if
\begin{equation}
    \valid_t \geq \target = N.
\end{equation}

\paragraph{Completion fidelity.}
An agent may report a count $\reported_t$ or claim completion. We define a \emph{false completion}
as a final action with a completion claim when $\valid_t < N$. We define \emph{premature stopping}
as an ask-user or non-completing final action before $\valid_t \geq N$. The normalized reported-count
error for a run with a reported count is
\begin{equation}
    \frac{|\reported - \valid|}{\max(1, N)}.
\end{equation}
Progress inflation occurs when $\reported > \valid$.

\paragraph{Repeated work.}
Because long-horizon agents often reprocess the same evidence or retry the same failed item, we
track repeated work separately from invalid work. In identifier-retrieval tasks, repeated work
appears as duplicate submissions. In backlog/verifier tasks, it appears as repeated attempts on an
already-passed unit or repeated failed attempts that do not change verified progress. This metric
captures wasted interaction budget and loss of external memory.

\paragraph{Verifier-gated completion.}
A controller is verifier-gated if it intercepts final or ask-user actions and refuses to terminate
the run while $\valid < N$. Verifier gating does not make the underlying policy more capable at
editing code, ranking evidence, or identifying relevant items. It changes the execution contract so
that unsupported completion claims cannot end the interaction.

\section{Benchmark Design}
\label{sec:benchmark}
\pushbench{} implements QGP through verifier-checkable task families in which each task specifies an
objective, a target count, a budget, and a verifier-derived success condition. Agents emit structured
actions, and outcomes are computed from verifier-visible state rather than final-message claims.
The workflow in Figure~\ref{fig:pushbench-overview} separates task definition, action execution,
verification, and metric computation.
Because every accepted item is recorded by the verifier, the benchmark can separate target
completion from false completion, duplicate submission, reported-count error, and premature stopping. 

The controlled task families
intentionally keep unit difficulty low-to-medium: if the units themselves
require open-ended coding or domain expertise, a failure may reflect unsolved work rather
than lost quantitative persistence. This design reduces that confound so repeated work, stale
progress state, and unsupported completion claims can be attributed directly to QGP failures.

\paragraph{\reposcan{} task family.}
\reposcan{} is a repository-scanning benchmark family. In our evaluation, task instances are
generated from local snapshots of \texttt{requests}, \texttt{pytest}, and \texttt{flask}
\citep{requestsRepo,pytestRepo,flaskRepo}. During an interaction, the agent searches
repository-derived candidates and submits stable artifact identifiers. A task succeeds only when the
hidden verifier accepts at least $N$ distinct submitted identifiers. Thus \reposcan{} tests whether
an agent maintains verifier-visible progress across many search-and-submit steps.

\paragraph{\dataops{} task family.}
\dataops{} extends QGP from identifier retrieval to verifier-backed work units. Each task instance receives a
backlog drawn from public data snippets, including Vega cars data \citep{vegaCarsData} and
FiveThirtyEight airline-safety data \citep{fivethirtyeightAirlineSafety}, plus
repository-derived fixtures built from the same \texttt{requests}, \texttt{pytest}, and
\texttt{flask} snapshots used by \reposcan{}. These fixtures turn repository files and metadata
into small checker-backed units, such as CSV-field checks, file-metadata repairs, or consistency
answers. A unit counts only after its checker accepts it, and the task succeeds only when at least
$N$ distinct units pass.

The two families are complementary. \reposcan{} isolates persistence in a search-and-submit setting
where valid work is a distinct repository identifier. \dataops{} keeps the same count invariant but
adds short work loops: inspecting an artifact, producing or editing an answer, and
submitting the unit only after a checker accepts it. This progression checks whether QGP
failures persist when the agent must do small pieces of work rather than only retrieve identifiers.

\paragraph{Policies and harnesses.}
\pushbench{} separates the policy that proposes work from the controller that executes or audits it.
Native LLMPolicy and LangGraph-backed policies are controlled execution backends rather than
representative samples of all agent architectures. They share task manifests, budgets, verifiers,
hidden valid sets, and summary fields, letting controller comparisons isolate changes in state
exposure, duplicate filtering, completion gating, and repair routing. The experimental setup below
specifies target counts, budgets, and aggregation units for each family. Additional task-family
construction details are in Appendix~\ref{app:benchmark-details}.

\section{Persistence Controllers}
\label{sec:controllers}
\pushbench{} distinguishes passive, verifier-gated, and stateful persistence controllers.
Across all modes, hidden valid identifiers and labels remain inside the verifier. Controllers
observe only the feedback produced for submitted units, such as accept/reject decisions, duplicate
status, verified counts, and remaining count.

\paragraph{\stateful{} for identifier retrieval.}
\stateful{} is a controller-level mechanism for search-and-submit tasks. The policy still chooses
actions, but the controller stores submitted identifiers, seen search pages, the last query, and the
next unseen page for each query. When the policy repeats a search page, the controller advances to
the next unseen page. When the policy submits identifiers, the controller filters
identifiers that were already submitted or repeated within the same action. If filtering leaves an
empty submission, the controller can repair the action into a search for the next unseen page. When
the policy asks the user or claims completion before $\valid \geq N$, the controller blocks the
termination and returns a verifier observation explaining that the target has not yet been met.

\paragraph{Standard and verifier-gated controls.}
The standard controller executes parsed policy actions without duplicate-aware persistence repair.
The verifier-gated controller adds completion gating: it blocks unsupported final or ask-user
termination while $\valid < N$. This isolates whether preventing false completion is sufficient
without adding stronger state over already-submitted artifacts.

\paragraph{\workunit{} for verifier-backed units.}
\workunit{} extends the same idea to backlog tasks. It tracks unit status, notices stale inspection
or no-submit loops, steers the policy back to the first pending or attempted unit, repairs
post-edit behavior toward verifier execution, and records verifier recoveries when a later repair
turns an initially failed or incomplete trajectory into a passing unit. As with \stateful{}, the
reported behavior includes visible controller assistance.

These controllers make a narrow claim: external state can enforce persistence invariants that are difficult to leave entirely to model memory. \pushbench{} reports passive and controller-assisted execution separately to keep this assistance visible.

\section{Experimental Setup}
\label{sec:setup}
Experiment 1 evaluates \reposcan{}. The \reposcan{} manifest has 36 task instances generated
from local snapshots of \texttt{requests}, \texttt{pytest}, and \texttt{flask}. Target counts are
$N \in \{10,25,50,100\}$, with nine task instances per target. Predicate families include
keyword-or-pattern, path-and-content, and test-or-documentation artifacts. The success invariant is
$\valid \geq \target$ over unique repository artifact identifiers.

Experiment 2 evaluates \dataops{}. Each task instance uses a backlog of public-data and repository-derived fixture
work units with unit verifiers. Targets are $N \in \{3,5,10,20\}$, with six backlogs
per target. The evaluation crosses two agent implementations, three controllers, four target counts, and
six seeds or backlogs per target. Success again requires
$\valid \geq \target$, now over verifier-passing work units rather than retrieved artifact IDs.

We use \emph{task instance} to mean one fixed manifest or backlog, and \emph{run} to mean one
policy-controller execution on one task instance. Thus a \reposcan{} row with 36 executions covers four
target counts with nine task instances each, while a \dataops{} cell with 24 executions covers four target
counts with six backlogs each. These controlled, verifier-checkable tasks are chosen to isolate
persistence from the confound that an agent may simply be unable to solve an open-ended unit.

The task manifests are constructed before evaluation and define the repository, target count,
budget, objective text, and hidden verifier state. In \reposcan{}, valid identifiers are generated
offline from stable repository artifact IDs under the three predicate families; agents receive the
objective and search observations, but not the hidden valid-ID set. In \dataops{}, each backlog unit
contains a public prompt, artifact path, and deterministic checker command; a unit counts only after
the checker accepts it. Budgets are fixed by target: \reposcan{} uses 30, 60, 100, and 180 steps for
$N=10,25,50,100$, respectively, while \dataops{} uses 30, 50, 90, and 160 steps for
$N=3,5,10,20$. Allowed actions are JSON-structured search, submit, final, and ask-user actions for
retrieval; work-unit tasks additionally allow inspection, edit, checker execution, and unit
submission.

\paragraph{Agents and models.}
For \reposcan{}, we evaluate three GPT-family models: \texttt{gpt-4.1-mini}, \texttt{gpt-4.1}, and
\texttt{gpt-5.4}. Native LLMPolicy and LangGraph are each evaluated under standard, verifier-gated, and
\stateful{} controllers, giving 36 task instances with one execution per task instance and controller condition, or 108
executions per backend-model pair
and 648 Native/LangGraph executions over fixed task instances. We also evaluate \letta{} and LangGraph+Memory under the standard
controller only, because they represent external memory baselines rather than \pushbench{}
controller ablations, adding two baselines $\times$ 36 task instances $\times$ 3 models = 216 executions. For
\dataops{}, we use the same three-model parity design:
Native LLMPolicy and LangGraph use standard, verifier-gated, and \workunit{} controllers, while
\letta{} and LangGraph+Memory use the standard controller as external memory baselines. Native and
LangGraph are controlled execution backends rather than representative samples of commercial or
open-source agent designs.
Additional coding-style and frontier-agent studies are discussed in Section~\ref{sec:discussion}
and detailed in the appendix.

\paragraph{Evaluation setup.}
All executions use fixed task manifests, fixed target-dependent budgets, and verifier-derived success
labels. Within each task family, controller comparisons use the same task, budget, verifier, and
valid set; only the controller's exposure of progress state, duplicate filtering, completion
gating, and repair routing changes. Detailed implementation artifacts and reliability metrics
are included for reproducibility.

\paragraph{Metrics.}
We report target success rate, average verified valid count, duplicate submit rate, and valid
artifacts per step. Success needs $\valid \geq \target$ and is evaluated from verifier state
not from the agent's final message.

\section{Experiment 1: Repository-Scanning Results}
\label{sec:results}
Table~\ref{tab:reposcan-main} reports the \reposcan{} results. Each row aggregates one execution on each
of 36 fixed task instances: nine task instances at each target count $N \in \{10,25,50,100\}$. Appendix
Figure~\ref{fig:target-scaling-appendix} decomposes these aggregate rows by target count.

\begin{table*}[t]
\centering
\scriptsize
\setlength{\tabcolsep}{3pt}
\resizebox{\textwidth}{!}{%
\begin{tabular}{lllrrrrr}
\toprule
Model & Agent & Controller & Task inst. & Succ. & Avg. valid & Dup. rate & Valid/step \\
\midrule
\texttt{gpt-4.1-mini} & Native & Standard & 36 & 0.028 & 6.42 & 0.880 & 0.103 \\
\texttt{gpt-4.1-mini} & Native & Verifier-gated & 36 & 0.083 & 6.64 & 0.872 & 0.128 \\
\texttt{gpt-4.1-mini} & Native & \methodhi{\stateful{}} & 36 & 0.722 & 38.67 & 0.000 & 2.229 \\
\texttt{gpt-4.1-mini} & LangGraph & Standard & 36 & 0.083 & 5.36 & 0.868 & 0.106 \\
\texttt{gpt-4.1-mini} & LangGraph & Verifier-gated & 36 & 0.028 & 6.44 & 0.881 & 0.103 \\
\texttt{gpt-4.1-mini} & LangGraph & \methodhi{\stateful{}} & 36 & 0.722 & 37.94 & 0.000 & 2.127 \\
\texttt{gpt-4.1-mini} & \letta{} & Standard & 36 & 0.000 & 2.39 & 0.129 & 0.179 \\
\texttt{gpt-4.1-mini} & LG+Memory & Standard & 36 & 0.056 & 2.25 & 0.468 & 0.389 \\
\midrule
\texttt{gpt-4.1} & Native & Standard & 36 & 0.306 & 17.11 & 0.557 & 0.615 \\
\texttt{gpt-4.1} & Native & Verifier-gated & 36 & 0.333 & 18.64 & 0.640 & 0.692 \\
\texttt{gpt-4.1} & Native & \methodhi{\stateful{}} & 36 & 0.722 & 38.53 & 0.000 & 2.276 \\
\texttt{gpt-4.1} & LangGraph & Standard & 36 & 0.306 & 17.78 & 0.681 & 0.700 \\
\texttt{gpt-4.1} & LangGraph & Verifier-gated & 36 & 0.389 & 19.33 & 0.747 & 0.589 \\
\texttt{gpt-4.1} & LangGraph & \methodhi{\stateful{}} & 36 & 0.722 & 38.92 & 0.000 & 2.328 \\
\texttt{gpt-4.1} & \letta{} & Standard & 36 & 0.444 & 31.83 & 0.033 & 0.412 \\
\texttt{gpt-4.1} & LG+Memory & Standard & 36 & 0.472 & 31.86 & 0.006 & 3.117 \\
\midrule
\texttt{gpt-5.4} & Native & Standard & 36 & 0.306 & 20.89 & 0.112 & 1.177 \\
\texttt{gpt-5.4} & Native & Verifier-gated & 36 & 0.472 & 26.47 & 0.216 & 0.804 \\
\texttt{gpt-5.4} & Native & \methodhi{\stateful{}} & 36 & 0.694 & 37.00 & 0.000 & 2.337 \\
\texttt{gpt-5.4} & LangGraph & Standard & 36 & 0.167 & 11.83 & 0.663 & 0.480 \\
\texttt{gpt-5.4} & LangGraph & Verifier-gated & 36 & 0.222 & 14.81 & 0.743 & 0.374 \\
\texttt{gpt-5.4} & LangGraph & \methodhi{\stateful{}} & 36 & 0.778 & 41.75 & 0.000 & 2.537 \\
\texttt{gpt-5.4} & \letta{} & Standard & 36 & 0.722 & 37.25 & 0.009 & 1.718 \\
\texttt{gpt-5.4} & LG+Memory & Standard & 36 & 0.722 & 36.69 & 0.005 & 3.353 \\
\bottomrule
\end{tabular}
}
\caption{\reposcan{} results. Each row aggregates 36 fixed real-project artifact-finding task
instances over \texttt{requests}, \texttt{pytest}, and \texttt{flask}: nine task instances at each target
count $10$, $25$, $50$, and $100$. Appendix Figure~\ref{fig:target-scaling-appendix} reports the
per-target breakdown. Native and LangGraph rows are matched controller comparisons; \letta{} and
LG+Memory are standard-controller external memory baselines. All values are from
verifier-visible state.}
\label{tab:reposcan-main}
\end{table*}

\paragraph{Main finding.}
Standard and verifier-gated agents improve with stronger models, showing that \reposcan{} is not 
impossible for non-persistent controllers. However, the same QGP failure mode remains visible:
standard and verifier-gated rows still degrade sharply on larger targets and retain duplicate or
premature-stop behavior. For example, \texttt{gpt-5.4} Native standard reaches 30.6\% success and a
much lower duplicate rate than smaller models, but \texttt{gpt-5.4} LangGraph standard reaches only
16.7\% success with a 0.663 duplicate submit rate. Verifier gating blocks unsupported termination,
yet it does not add the duplicate-aware progress state required to make large target counts reliable.

By contrast, \stateful{} changes the execution profile across all three models and both agent implementations.
It reaches 72.2\% success for \texttt{gpt-4.1-mini} with both Native and LangGraph, 72.2\% for
\texttt{gpt-4.1} with both backends, 69.4\% for \texttt{gpt-5.4} Native, and 77.8\% for
\texttt{gpt-5.4} LangGraph. It also keeps duplicate submit rate at 0.000 in every row. Because
tasks, budgets, verifiers, and valid sets are matched, this improvement isolates the value of
controller-visible progress state rather than a change in the underlying task.

\paragraph{Target scaling.}
The target-count breakdown shows that small-target success does not imply persistence at larger
targets. Many standard and verifier-gated rows succeed at $N=10$, but success often drops to zero by
$N=50$ or $N=100$. \stateful{} remains strongest at small and medium targets and preserves zero
duplicate submissions across all targets, although high-target conditions can still exhaust the budget.
This is the intended interpretation: controller persistence reduces repeated work and premature
completion, but it does not make every long-horizon retrieval target free.
Appendix~\ref{app:target-scaling}, Figure~\ref{fig:target-scaling-appendix}, reports the same
pattern against memory baselines: memory can reduce duplicates, but its high-target success is
model-dependent, while \stateful{} preserves the most consistent target scaling.

\paragraph{External memory baselines.}
\letta{} and LangGraph+Memory test whether run-local memory is enough to maintain progress without a QGP-specific controller. 
The results show that it helps for stronger models, but the effect is not consistent across models. 
With \texttt{gpt-4.1-mini}, both memory baselines remain far below
\stateful{}: \letta{} completes 0 of 36 task instances and LangGraph+Memory completes 2 of 36, while \stateful{} completes
72.2\% for both Native and LangGraph. With \texttt{gpt-4.1}, memory improves substantially:
\letta{} reaches 44.4\% and LangGraph+Memory reaches 47.2\%, but both remain below the 72.2\%
\stateful{} rows. With \texttt{gpt-5.4}, \letta{} and LangGraph+Memory both reach 72.2\%,
competitive with Native \stateful{} at 69.4\% and close to LangGraph \stateful{} at 77.8\%. Thus
memory can help with stronger policies and sometimes approaches controller-level persistence, but
it is not a uniform replacement across model scales. Appendix~\ref{app:langgraph-memory} reports
the detailed breakdown.

\section{Experiment 2: Work-Unit Results}
\label{sec:dataops-results}
\dataops{} asks whether the same persistence pattern survives when units are no longer just
retrieved identifiers. \reposcan{} isolates counting and retrieval over repository artifacts;
\dataops{} tests whether the same state problem appears when each unit must be inspected, checked,
edited or answered, verified, and submitted. \dataops{} uses public data snippets and repository-derived
fixtures, but still keeps unit difficulty low enough that the experiment is mainly about
persistence, state fidelity, and recovery rather than open-ended software engineering.

\begin{figure*}[t]
\centering
\includegraphics[width=\textwidth]{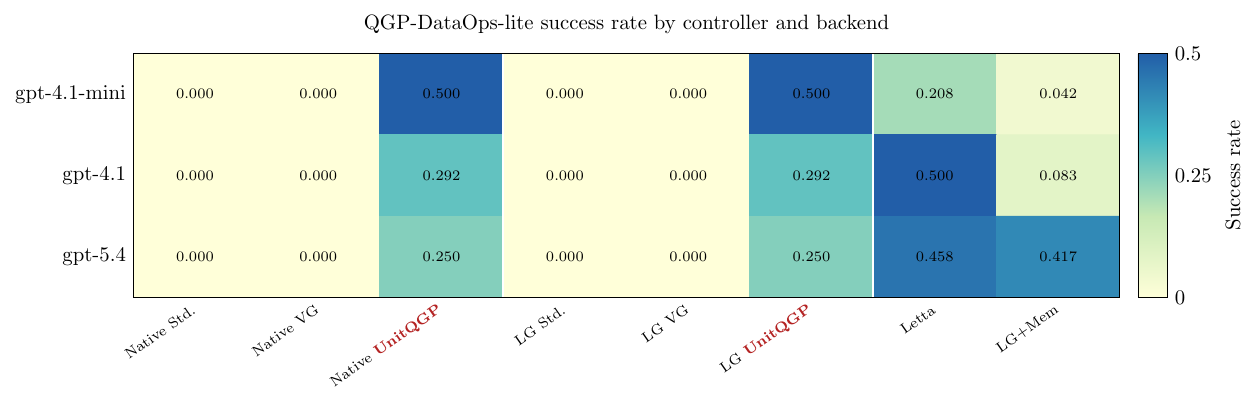}
\caption{\dataops{} success rates. Native and LangGraph (LG) use standard, verifier-gated (VG),
and \workunit{} controllers. \letta{} and LangGraph+Memory (LG+Mem) are standard-controller memory
baselines. The heatmap emphasizes that passive and completion-gated controllers complete no task instances,
while backlog-tracking and memory-based variants recover nonzero completion.}
\label{fig:dataops-main}
\end{figure*}

Figure~\ref{fig:dataops-main} reports the \dataops{} comparison. Each cell aggregates one execution on
each of 24 fixed backlogs for one model: six backlogs at each target count $N \in \{3,5,10,20\}$ for each
controller or memory baseline.

\paragraph{Main finding.}
Standard and verifier-gated controllers finish no \dataops{} task instances across all three models and both
controlled backends. This strengthens the work-unit conclusion: once tasks require inspection,
checking, small edits or answers, and verifier-backed submission, stronger policies alone do not
solve the quantitative stopping problem under passive execution. 
Verifier gating prevents unsupported completion, but lacks the progress state needed to finish the backlog.

\workunit{} restores nonzero completion in every model-backend pair: 50.0\% for
\texttt{gpt-4.1-mini}, 29.2\% for \texttt{gpt-4.1}, and 25.0\% for \texttt{gpt-5.4} under both
Native and LangGraph. The absolute success rates are lower than \reposcan{}, as expected for
verifier-backed work units, but the controller contrast is cleaner: the matched standard and
verifier-gated rows are all zero.

\paragraph{Memory baselines.}
Memory baselines are useful but model-dependent. \letta{} reaches 20.8\%, 50.0\%, and 45.8\%
success from \texttt{gpt-4.1-mini} to \texttt{gpt-5.4}, while LangGraph+Memory reaches 4.2\%,
8.3\%, and 41.7\%. This mirrors the \reposcan{} pattern: generic memory can be competitive with a
strong policy, but it is not a uniform substitute for verifier-aligned progress state. 
We make the work-unit conclusion conservatively: controller-level \workunit{} is one effective
implementation of verifier-aligned state, while sufficiently strong memory systems can sometimes
recover comparable behavior.

\paragraph{Failure analysis.}
Inspection of representative executions exposes several distinct failure modes. Some standard executions claim
completion with more units than the verifier accepts, while others stop early or exhaust budget
without making enough accepted submissions. Verifier gating addresses unsupported final claims but
not no-submit or stale-inspection loops. In \workunit{} executions, the controller repairs stale
inspection, missing verifier calls, and no-submit loops by steering the agent back toward pending
units and checks. These repairs are controller-visible interventions, not hidden model
improvements, and executions that hit conservative no-progress limits are still classified as
budget-exhausted. Appendix~\ref{app:dataops-details} reports full controller metrics, paired
controller deltas, and a unit-kind breakdown for this evaluation.

\section{Discussion and Additional Evaluations}
\label{sec:discussion}
\paragraph{QGP as operational reliability.}
The experiments show that many failures are not zero-progress failures. Agents often find or
complete valid units, but lose the run by repeating accepted work, stopping before the target count,
or failing to use verifier feedback to choose the next unit. QGP makes this gap measurable: success
requires accumulated verifier-accepted progress, not a plausible final claim or a handful of correct
local steps. This is why the same failure pattern appears in both \reposcan{} and
\dataops{}.

\paragraph{Controllers as visible interventions.}
The controller comparisons show that progress state is not just bookkeeping around the agent; it
changes the trajectory. Passive controllers leave the policy to remember which units were submitted,
which were accepted, and whether final completion is justified. State-tracking controllers expose
that state at each step and use it to block unsupported final actions, filter duplicates, advance
repeated searches, and route work-unit agents back to pending checks. Since the verifier, task
instances, and budgets are fixed, the gains point to a runtime failure mode: without visible
progress state, agents can spend the budget repeating work they have done or ending
before the count invariant is met. Appendix~\ref{app:stateful-ablation} decomposes the \stateful{}
components.

\paragraph{Memory and coding-style evaluations.}
The extra evaluations enhance the same conclusion under less controlled conditions: QGP
failures are reduced by stronger infrastructure, but they do not disappear as a measurement
problem. 
The memory baselines show that generic run-local memory can help, especially with
stronger policies, yet the gains are model-dependent and do not uniformly replace verifier-aligned
progress state. In the coding-style \swelite{} evaluation, passive and verifier-gated
Native/LangGraph agents complete no task instances, while \workunit{} recovers 37.5--41.7\%
success. Aider \citep{aiderRepo} succeeds on most task instances, indicating that the units are feasible, but
also that coding-agent harness behavior changes the problem being measured. Appendix~\ref{app:swelite-results}
reports the detailed breakdown.

\paragraph{Frontier agents.}
The black-box evaluation provides a complementary check with stronger coding-agent products: they solve many smaller-count cases, but the larger explicit-count goal still exposes persistence failures.
 Claude Code with Sonnet 4.6 and Codex CLI with \texttt{gpt-5.4} solve many
50-artifact \reposcan{} cases, but each condition drops to 3/9 successes at 100 artifacts.
Adding an explicit checklist prompt does not improve paired success, suggesting that generic
reminders are weaker than verifier-aligned progress tracking. Appendix~\ref{app:commercial-reposcan}
and Figure~\ref{fig:commercial-reposcan-summary} report the details.

\section{Conclusion}
We introduced \qgp{}, a formulation for evaluating whether long-horizon agents persist until an
explicit quantitative completion invariant is externally satisfied. 
Our experiments suggest that high-count persistence
remains nontrivial even for stronger black-box frontier-agent harnesses. 
We hope QGP-style metrics become a standard part of agent evaluation, complementing end-to-end task success with a direct measure of whether agents preserve verified progress until the requested work is actually complete.

\section{Limitations}
This work has five main limitations.

\begin{enumerate}
	\item First, \reposcan{} and \dataops{} are controlled QGP benchmarks, not general software-agent
leaderboards. The included models test whether persistence failures remain visible under stronger
policies. They are designed to isolate quantitative persistence under controlled conditions, not to rank general-purpose agents or estimate their overall software-engineering ability.
    \item  Second, the task coverage is still limited. \reposcan{} uses three repositories and 36 task
instances, while \dataops{} uses public data snippets and repository-derived fixtures. Broader
repositories, data sources, and task types are needed before making claims about software or
data-agent workloads in general.
\item Third, the external memory baselines are practical comparisons rather than controlled ablations.
Systems such as \letta{} and LangGraph+Memory include their own runtime and memory semantics, so
their results should be read as evidence about whether generic memory helps, not as isolated tests
of a single controller component.
\item Fourth, the coding-style and black-box frontier-agent evaluations trade control for realism.
Coding tasks mix persistence with coding skill, repository navigation, editing ability, and harness
behavior. Frontier-agent products add hidden prompting, memory, tool routing, retry behavior, and
product updates. These evaluations therefore provide external-validity evidence, but they do not
establish general weaknesses of commercial or open-source agents.
\item Finally, \stateful{} and \workunit{} require an online verifier or checker. The controller does not
need access to the hidden valid set, but it does need feedback on submitted units during execution.
The approach does not directly apply to tasks where progress can only be judged at the end by a
human. When explicit count goals are only one part of success, QGP should also be complemented with
qualitative evaluation.
\end{enumerate}

\section{Ethical Considerations}
\pushbench{} is intended as an evaluation framework for agent reliability. The main
ethical risk is overinterpreting small controlled pilots as general claims about model safety or
competence. We mitigate this by making controller assistance explicit and by reserving numeric
claims for verifier-derived metrics. Software repository and public-data sources are cited in the
benchmark descriptions.

\section{Open Data and Artifacts}
An anonymized \pushbench{} artifact is available at
\url{https://anonymous.4open.science/r/artifacts-pushbench-7DB1/README.md}. It includes the code,
manifests, repository snapshots, aggregate summaries, table data, and reproduction scripts needed
to regenerate the reported tables and run verifier smoke checks. The artifact excludes raw model
traces, private endpoint configuration, and non-anonymized local paths.

\bibliography{custom}

\appendix

The appendix gives more details of the evaluation. Appendix~\ref{app:benchmark-details}
gives task-construction details for \pushbench{}. Appendix~\ref{app:reposcan-analyses} reports
additional \reposcan{} memory, target-scaling, and ablation analyses. Appendix~\ref{app:dataops-details}
expands the \dataops{} results. Appendix~\ref{app:external-validity} contains coding-style and
frontier-agent evaluations. All appendix results use the same verifier-derived metrics as the
experiments above.

\section{PushBench Benchmark Details}
\label{app:benchmark-details}
\reposcan{} task instances vary by repository, target count, seed, and predicate family. Artifacts
may be source, test, documentation, or configuration items, and predicate families include
keyword-or-pattern, path-and-content, and test-or-documentation membership. Agents see the objective
and search observations, but the valid identifier set remains hidden inside the verifier.

\dataops{} backlogs combine public data snippets and repository-derived fixtures. Public snippets come
from Vega cars data \citep{vegaCarsData} and FiveThirtyEight airline-safety data
\citep{fivethirtyeightAirlineSafety}; repository fixtures are derived from the same local
\texttt{requests}, \texttt{pytest}, and \texttt{flask} snapshots used by \reposcan{}. They are
small, deterministic tasks over repository files or metadata rather than open-ended code changes.
Units include CSV schema or count checks, consistency-check answers, metadata repairs,
deterministic verifier execution, and submission of the accepted unit. These units are intended to
require concrete checking and recovery without making the underlying work open-ended.

\paragraph{Additional task variants.}
The same search-submit-verify interface can be instantiated for synthetic retrieval corpora and
code-location retrieval over repository text chunks. 
The reported claims use the \reposcan{} and \dataops{}
experiments, plus the coding-style and frontier-agent evaluations in
Appendix~\ref{app:external-validity}.

\section{Additional RepoScan Analyses}
\label{app:reposcan-analyses}

\subsection{Memory Baseline Breakdown}
\label{app:langgraph-memory}
Table~\ref{tab:langgraph-memory-appendix} expands the memory rows in Table~\ref{tab:reposcan-main} with failure-mode counts that are not shown in the main result table.
The
LangGraph+Memory baseline uses the same 36 \reposcan{} tasks and budgets, uses only the standard
controller, and records previous compact observations, status snapshots, actions, submitted
identifiers, searched pages, and parse errors. 
It does not receive hidden valid identifiers or
verifier labels.

\begin{table*}[t]
\centering
\scriptsize
\setlength{\tabcolsep}{4pt}
\resizebox{\textwidth}{!}{%
\begin{tabular}{llrrrrrr}
\toprule
Model & Baseline & Succ. & Avg. valid & Dup. rate & Premature & Budget exh. & Valid/step \\
\midrule
\texttt{gpt-4.1-mini} & LangGraph Standard & 0.083 & 5.36 & 0.918 & 0.000 & 0.917 & 0.060 \\
\texttt{gpt-4.1-mini} & LangGraph+Memory & 0.056 & 2.25 & 0.468 & 0.583 & 0.000 & 0.389 \\
\texttt{gpt-4.1-mini} & LangGraph \methodhi{\stateful{}} & 0.722 & 37.94 & 0.000 & 0.000 & 0.278 & 0.781 \\
\texttt{gpt-4.1-mini} & \letta{} Standard & 0.000 & 2.39 & 0.159 & 0.889 & 0.111 & 0.169 \\
\midrule
\texttt{gpt-4.1} & LangGraph Standard & 0.306 & 17.78 & 0.858 & 0.083 & 0.611 & 0.385 \\
\texttt{gpt-4.1} & LangGraph+Memory & 0.472 & 31.86 & 0.006 & 0.111 & 0.000 & 3.117 \\
\texttt{gpt-4.1} & LangGraph \methodhi{\stateful{}} & 0.722 & 38.92 & 0.000 & 0.000 & 0.278 & 0.820 \\
\texttt{gpt-4.1} & \letta{} Standard & 0.444 & 31.83 & 0.062 & 0.111 & 0.444 & 0.392 \\
\midrule
\texttt{gpt-5.4} & LangGraph Standard & 0.167 & 11.83 & 0.867 & 0.583 & 0.250 & 0.274 \\
\texttt{gpt-5.4} & LangGraph+Memory & 0.722 & 36.69 & 0.005 & 0.278 & 0.000 & 3.353 \\
\texttt{gpt-5.4} & LangGraph \methodhi{\stateful{}} & 0.778 & 41.75 & 0.000 & 0.000 & 0.222 & 0.977 \\
\texttt{gpt-5.4} & \letta{} Standard & 0.722 & 37.25 & 0.017 & 0.278 & 0.000 & 1.492 \\
\bottomrule
\end{tabular}
}
\caption{\reposcan{} memory baseline breakdown. Generic memory can sharply reduce duplicate
submissions, especially with stronger models, but success and stopping behavior remain
model-dependent.}
\label{tab:langgraph-memory-appendix}
\end{table*}

LangGraph+Memory is model-dependent. With \texttt{gpt-4.1-mini}, it does not improve success over
plain LangGraph and remains far below \stateful{}. With \texttt{gpt-4.1}, it improves standard
LangGraph from 30.6\% to 47.2\% success and nearly eliminates duplicate submissions, but still
trails LangGraph \stateful{} at 72.2\%. With \texttt{gpt-5.4}, LangGraph+Memory reaches 72.2\%,
matching \letta{} and approaching LangGraph \stateful{} at 77.8\%. The pattern refines the memory
comparison: memory can be a viable implementation strategy when paired with a strong policy, while
controller-level persistence remains a simple and consistently strong way to expose and enforce the
QGP state contract.

\subsection{Target Scaling and Paired Uncertainty}
\label{app:target-scaling}
\begin{figure*}[t]
\centering
\includegraphics[width=\textwidth]{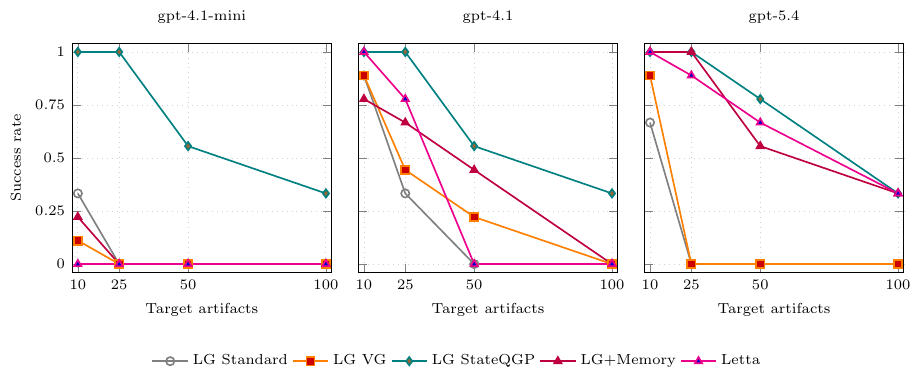}
\caption{\reposcan{} target scaling. Each point aggregates the nine task instances in one target
bucket from Table~\ref{tab:reposcan-main}. Many baselines solve small targets but collapse as the
requested artifact count grows; \stateful{} preserves nonzero high-target success while maintaining
zero duplicate submissions.}
\label{fig:target-scaling-appendix}
\end{figure*}

Figure~\ref{fig:target-scaling-appendix} decomposes the aggregate \reposcan{} rows from
Table~\ref{tab:reposcan-main} by target count for LangGraph, memory baselines, and \stateful{}.
Each target bucket contains the same nine fixed task instances used in Table~\ref{tab:reposcan-main}. The central
pattern is target scaling rather than small-target success: many rows solve $N=10$ but collapse by
$N=50$ or $N=100$, while \stateful{} remains nonzero at the largest target for all three models.

A paired bootstrap over matched task instances gives the same picture. For each bootstrap sample,
we resample task instances with replacement and recompute the success-rate difference between two
controllers on the same sampled tasks. The reported interval is therefore an uncertainty range for
the difference, not for the success rate itself. On \texttt{gpt-4.1} LangGraph, \stateful{} raises
success over the standard controller by 41.7 percentage points, with a 95\% interval from 27.8 to
58.3 points. Verifier gating raises success by only 8.3 points, and its interval ranges from -2.8 to
22.2 points; the negative lower end means that, under some resampled task sets, verifier gating
does not improve over the standard controller. For \texttt{gpt-5.4} LangGraph, the \stateful{} gain
over standard is 61.1 points, with a 95\% interval from 44.4 to 77.8 points.

\subsection{Stateful Component Ablation}
\label{app:stateful-ablation}
\begin{figure*}[t]
\centering
\includegraphics[width=\textwidth]{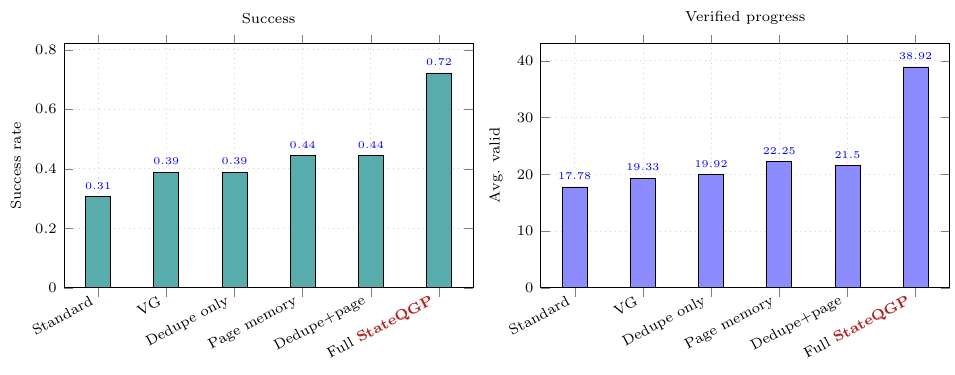}
\caption{Focused \texttt{gpt-4.1} LangGraph stateful ablation. Success and average verified count
rise most sharply only when duplicate filtering, page memory, and buffered verifier-aligned
progress state are combined in full \stateful{}.}
\label{fig:stateful-ablation}
\end{figure*}

We use a focused \texttt{gpt-4.1} LangGraph ablation to separate the main components of
\stateful{}. The ablation reuses the same 36 \reposcan{} tasks and adds three variants: duplicate
filtering only, repeated-search page memory only, and duplicate filtering plus page memory without
buffered seen-document submission. Figure~\ref{fig:stateful-ablation} compares these variants
against the standard, verifier-gated, and full \stateful{} rows, which are the
matched \reposcan{} rows for the same model and backend.

The ablation shows that no single passive memory feature explains the \stateful{} result. Duplicate
filtering eliminates duplicate submissions but does not improve success beyond verifier gating.
Page memory improves success to 44.4\%, but the full controller reaches 72.2\%. The paired success
delta between full \stateful{} and the no-buffer dedupe+page variant is 0.278 with a 95\% bootstrap
interval of [0.139, 0.417]. The key mechanism is therefore not generic memory alone, but
verifier-aligned progress state that can turn seen candidates into forward progress.

\section{DataOps-lite Detailed Results}
\label{app:dataops-details}
Table~\ref{tab:dataops-full} reports the full \dataops{} controller metrics behind
Figure~\ref{fig:dataops-main}. Standard and verifier-gated failures are dominated by budget
exhaustion, premature stopping, and very low accepted-unit counts. 
Table~\ref{tab:dataops-paired} gives paired
controller deltas over matched tasks. \workunit{} improves success over both standard and
verifier-gated controllers for every model-backend pair, while verifier gating alone has zero
success delta over standard. Table~\ref{tab:dataops-unit-kind} groups executions by unit kind. \workunit{}
is strongest on artifact-validation units and weaker on consistency-check and metadata-update
units; memory-baseline successes concentrate in mixed backlogs, which helps rule out a single
retrieval-style shortcut.

\begin{table*}[t]
\centering
\scriptsize
\setlength{\tabcolsep}{2pt}
\resizebox{\textwidth}{!}{%
\begin{tabular}{lllrrrrrrrr}
\toprule
Model & Backend & Controller & Backlogs & Succ. & Avg. valid & Dup. rate & Valid/step &
Budget exh. & Premature & False comp. \\
\midrule
\texttt{gpt-4.1-mini} & Native & Standard & 24 & 0.000 & 0.708 & 0.125 & 0.036 & 0.833 & 0.167 & 0.000 \\
\texttt{gpt-4.1-mini} & Native & VG & 24 & 0.000 & 0.833 & 0.083 & 0.037 & 1.000 & 0.000 & 0.000 \\
\texttt{gpt-4.1-mini} & Native & \methodhi{\workunit{}} & 24 & 0.500 & 8.000 & 0.059 & 0.136 & 0.500 & 0.000 & 0.000 \\
\texttt{gpt-4.1-mini} & LangGraph & Standard & 24 & 0.000 & 0.708 & 0.050 & 0.046 & 0.917 & 0.083 & 0.000 \\
\texttt{gpt-4.1-mini} & LangGraph & VG & 24 & 0.000 & 0.792 & 0.042 & 0.042 & 1.000 & 0.000 & 0.000 \\
\texttt{gpt-4.1-mini} & LangGraph & \methodhi{\workunit{}} & 24 & 0.500 & 8.042 & 0.101 & 0.144 & 0.500 & 0.000 & 0.000 \\
\texttt{gpt-4.1-mini} & \letta{} & Standard & 24 & 0.208 & 4.917 & 0.044 & 0.242 & 0.000 & 0.792 & 0.000 \\
\texttt{gpt-4.1-mini} & LG+Mem & Standard & 24 & 0.042 & 2.250 & 0.253 & 0.054 & 0.208 & 0.583 & 0.167 \\
\midrule
\texttt{gpt-4.1} & Native & Standard & 24 & 0.000 & 0.875 & 0.483 & 0.046 & 0.792 & 0.208 & 0.000 \\
\texttt{gpt-4.1} & Native & VG & 24 & 0.000 & 2.417 & 0.216 & 0.089 & 1.000 & 0.000 & 0.000 \\
\texttt{gpt-4.1} & Native & \methodhi{\workunit{}} & 24 & 0.292 & 7.333 & 0.569 & 0.102 & 0.708 & 0.000 & 0.000 \\
\texttt{gpt-4.1} & LangGraph & Standard & 24 & 0.000 & 1.125 & 0.558 & 0.057 & 0.792 & 0.208 & 0.000 \\
\texttt{gpt-4.1} & LangGraph & VG & 24 & 0.000 & 2.625 & 0.427 & 0.069 & 1.000 & 0.000 & 0.000 \\
\texttt{gpt-4.1} & LangGraph & \methodhi{\workunit{}} & 24 & 0.292 & 7.167 & 0.604 & 0.098 & 0.708 & 0.000 & 0.000 \\
\texttt{gpt-4.1} & \letta{} & Standard & 24 & 0.500 & 7.500 & 0.049 & 0.245 & 0.000 & 0.500 & 0.000 \\
\texttt{gpt-4.1} & LG+Mem & Standard & 24 & 0.083 & 5.833 & 0.013 & 0.190 & 0.250 & 0.500 & 0.167 \\
\midrule
\texttt{gpt-5.4} & Native & Standard & 24 & 0.000 & 0.000 & 0.000 & 0.000 & 0.542 & 0.458 & 0.000 \\
\texttt{gpt-5.4} & Native & VG & 24 & 0.000 & 0.000 & 0.000 & 0.000 & 1.000 & 0.000 & 0.000 \\
\texttt{gpt-5.4} & Native & \methodhi{\workunit{}} & 24 & 0.250 & 6.708 & 0.000 & 0.094 & 0.750 & 0.000 & 0.000 \\
\texttt{gpt-5.4} & LangGraph & Standard & 24 & 0.000 & 0.000 & 0.000 & 0.000 & 1.000 & 0.000 & 0.000 \\
\texttt{gpt-5.4} & LangGraph & VG & 24 & 0.000 & 0.000 & 0.000 & 0.000 & 1.000 & 0.000 & 0.000 \\
\texttt{gpt-5.4} & LangGraph & \methodhi{\workunit{}} & 24 & 0.250 & 6.542 & 0.060 & 0.092 & 0.750 & 0.000 & 0.000 \\
\texttt{gpt-5.4} & \letta{} & Standard & 24 & 0.458 & 7.042 & 0.040 & 0.290 & 0.000 & 0.542 & 0.000 \\
\texttt{gpt-5.4} & LG+Mem & Standard & 24 & 0.417 & 7.208 & 0.018 & 0.299 & 0.375 & 0.208 & 0.000 \\
\bottomrule
\end{tabular}
}
\caption{\dataops{} full controller metrics. Each row aggregates 24 fixed backlogs, with six at
each target count. Provider-error rate is 0.000 for every row in the evaluation.}
\label{tab:dataops-full}
\end{table*}

\begin{table*}[t]
\centering
\scriptsize
\setlength{\tabcolsep}{4pt}
\resizebox{\textwidth}{!}{%
\begin{tabular}{lllrrrrrr}
\toprule
Model & Backend & Delta & Paired tasks & Succ. delta & 95\% CI & Avg-valid delta &
Left-only & Right-only \\
\midrule
\texttt{gpt-4.1-mini} & Native & \methodhi{\workunit{}}--Std. & 24 & 0.500 & [0.292, 0.708] & 7.292 & 12 & 0 \\
\texttt{gpt-4.1-mini} & Native & \methodhi{\workunit{}}--VG & 24 & 0.500 & [0.292, 0.708] & 7.167 & 12 & 0 \\
\texttt{gpt-4.1-mini} & Native & VG--Std. & 24 & 0.000 & [0.000, 0.000] & 0.125 & 0 & 0 \\
\texttt{gpt-4.1-mini} & LangGraph & \methodhi{\workunit{}}--Std. & 24 & 0.500 & [0.292, 0.708] & 7.333 & 12 & 0 \\
\texttt{gpt-4.1-mini} & LangGraph & \methodhi{\workunit{}}--VG & 24 & 0.500 & [0.292, 0.708] & 7.250 & 12 & 0 \\
\texttt{gpt-4.1-mini} & LangGraph & VG--Std. & 24 & 0.000 & [0.000, 0.000] & 0.083 & 0 & 0 \\
\texttt{gpt-4.1} & Native & \methodhi{\workunit{}}--Std. & 24 & 0.292 & [0.125, 0.458] & 6.458 & 7 & 0 \\
\texttt{gpt-4.1} & Native & \methodhi{\workunit{}}--VG & 24 & 0.292 & [0.125, 0.500] & 4.917 & 7 & 0 \\
\texttt{gpt-4.1} & Native & VG--Std. & 24 & 0.000 & [0.000, 0.000] & 1.542 & 0 & 0 \\
\texttt{gpt-4.1} & LangGraph & \methodhi{\workunit{}}--Std. & 24 & 0.292 & [0.125, 0.500] & 6.042 & 7 & 0 \\
\texttt{gpt-4.1} & LangGraph & \methodhi{\workunit{}}--VG & 24 & 0.292 & [0.125, 0.500] & 4.542 & 7 & 0 \\
\texttt{gpt-4.1} & LangGraph & VG--Std. & 24 & 0.000 & [0.000, 0.000] & 1.500 & 0 & 0 \\
\texttt{gpt-5.4} & Native & \methodhi{\workunit{}}--Std. & 24 & 0.250 & [0.083, 0.417] & 6.708 & 6 & 0 \\
\texttt{gpt-5.4} & Native & \methodhi{\workunit{}}--VG & 24 & 0.250 & [0.083, 0.458] & 6.708 & 6 & 0 \\
\texttt{gpt-5.4} & Native & VG--Std. & 24 & 0.000 & [0.000, 0.000] & 0.000 & 0 & 0 \\
\texttt{gpt-5.4} & LangGraph & \methodhi{\workunit{}}--Std. & 24 & 0.250 & [0.083, 0.417] & 6.542 & 6 & 0 \\
\texttt{gpt-5.4} & LangGraph & \methodhi{\workunit{}}--VG & 24 & 0.250 & [0.083, 0.417] & 6.542 & 6 & 0 \\
\texttt{gpt-5.4} & LangGraph & VG--Std. & 24 & 0.000 & [0.000, 0.000] & 0.000 & 0 & 0 \\
\bottomrule
\end{tabular}
}
\caption{\dataops{} paired controller deltas over matched tasks. Left-only counts tasks solved by
the left controller but not the right controller; right-only counts the converse. Memory baselines
are excluded from this causal controller comparison.}
\label{tab:dataops-paired}
\end{table*}

\begin{table*}[t]
\centering
\footnotesize
\setlength{\tabcolsep}{6pt}
\begin{tabular}{@{}lccc@{}}
\toprule
Unit kind & Std./VG & \methodhi{\workunit{}} & Memory \\
\midrule
Artifact val. & 36 / 0.000 / 0.00 & 18 / 1.000 / 3.67 & 18 / 0.000 / 0.00 \\
Consistency & 24 / 0.000 / 0.00 & 12 / 0.000 / 0.33 & 12 / 0.000 / 0.00 \\
Metadata & 12 / 0.000 / 0.00 & 6 / 0.167 / 2.17 & 6 / 0.000 / 0.00 \\
Mixed & 216 / 0.000 / 1.12 & 108 / 0.287 / 8.96 & 108 / 0.380 / 7.72 \\
\bottomrule
\end{tabular}
\caption{\dataops{} unit-kind/source breakdown. Each cell reports executions / success rate / average
valid units. All unit-kind rows have \texttt{data\_source=mixed}, so source breakdown is
equivalent to the unit-kind grouping shown here.}
\label{tab:dataops-unit-kind}
\end{table*}

\section{External-Validity Evaluations}
\label{app:external-validity}

\subsection{SWE-lite Realism Evaluation}
\label{app:swelite-results}
\swelite{} is a coding-style extension of \pushbench{} inspired by software-engineering agent
benchmarks such as SWE-bench and SWE-agent \citep{jimenez2024swebench,yang2024sweagent}. Instead
of asking the agent to retrieve identifiers, each valid unit is an issue-style repair or check that
must pass a local verifier. The units are constructed from isolated real-workspace copies of
\texttt{requests}, \texttt{flask}, and \texttt{pytest}
\citep{requestsRepo,flaskRepo,pytestRepo}. Targets are $N \in \{1,2,3,5\}$, with six repo/seed
conditions per target, for 192 completed executions across Native LLMPolicy, LangGraph, and
Aider~\citep{aiderRepo}. These results are a coding-style external-validity check because the
setting entangles persistence with coding skill, repository navigation, editing ability, and
harness behavior.

\begin{table*}[t]
\centering
\scriptsize
\setlength{\tabcolsep}{4pt}
\resizebox{\textwidth}{!}{%
\begin{tabular}{llrrrrrr}
\toprule
Agent & Controller & Conditions & Succ. & Avg. valid & Budget exhausted & Valid/step & Valid/submit \\
\midrule
Native & Standard & 24 & 0.000 & 0.00 & 1.000 & 0.000 & 0.000 \\
Native & Verifier-gated & 24 & 0.000 & 0.00 & 1.000 & 0.000 & 0.000 \\
Native & \methodhi{\workunit{}} & 24 & 0.375 & 1.25 & 0.625 & 0.022 & 0.250 \\
LangGraph & Standard & 24 & 0.000 & 0.00 & 1.000 & 0.000 & 0.000 \\
LangGraph & Verifier-gated & 24 & 0.000 & 0.00 & 1.000 & 0.000 & 0.000 \\
LangGraph & \methodhi{\workunit{}} & 24 & 0.417 & 1.46 & 0.583 & 0.027 & 0.294 \\
Aider & Standard & 24 & 0.917 & 2.67 & 0.083 & 0.173 & 0.653 \\
Aider & \methodhi{\workunit{}} & 24 & 0.917 & 2.67 & 0.083 & 0.176 & 0.667 \\
\bottomrule
\end{tabular}
}
\caption{\swelite{} realism evaluation results. The benchmark contains issue-style repair
units in \texttt{requests}, \texttt{flask}, and \texttt{pytest}. The table reports how QGP evaluation
behaves in coding-style workflows.}
\label{tab:swelite-main}
\end{table*}

Native and LangGraph standard and verifier-gated controllers complete no \swelite{} task instances. Verifier
gating again prevents unsupported completion from ending the run, but it does not repair empty
search loops, stale inspection, missing unit submissions, or failure to run the verifier after
edits. \workunit{} materially changes the outcome for both implementations, reaching 37.5\%
success for Native and 41.7\% for LangGraph, with average verified counts rising from 0.00 to
1.25 and 1.46 respectively.

Aider provides a black-box solvability contrast. Its 91.7\% success rate under both standard and
\workunit{} execution shows that the \swelite{} units are feasible under the model and verifier
budget. The contrast also illustrates why coding-agent harnesses are reported separately: their
editing loop changes more than the QGP controller.

\paragraph{Black-box agent caveat.}
Commercial or product-style coding agents can be evaluated with the same verifier-facing
submission interface. Hidden prompting, memory, tool routing, retry behavior, and product updates
make their internal behavior only partially observable, so these evaluations complement rather than
replace the controlled controller comparisons.

\subsection{Black-Box Frontier-Agent RepoScan Evaluation}
\label{app:commercial-reposcan}
To test whether stronger commercial or frontier agent products still expose QGP failures, we
evaluate \reposcan{} at the hardest target counts. The
evaluation uses the 18 \reposcan{} tasks with $N \in \{50,100\}$, two black-box harnesses, and two
prompt modes. The default prompt asks the agent to continue until the verifier status reaches the
target. The checklist prompt adds explicit instructions to track accepted identifiers, rejected or
duplicate identifiers, searched query/page pairs, remaining count, and final eligibility. Agents
interact only through a thin local tool exposing \texttt{search}, \texttt{submit}, \texttt{status},
and \texttt{final}; hidden valid identifiers are kept inside the verifier process.
Claude Code is invoked with \texttt{--model sonnet} and reported as Claude Code Sonnet 4.6
(\texttt{claude-sonnet-4-6}). Codex CLI is invoked with \texttt{-m gpt-5.4}.

\begin{figure*}[t]
\centering
\includegraphics[width=\textwidth]{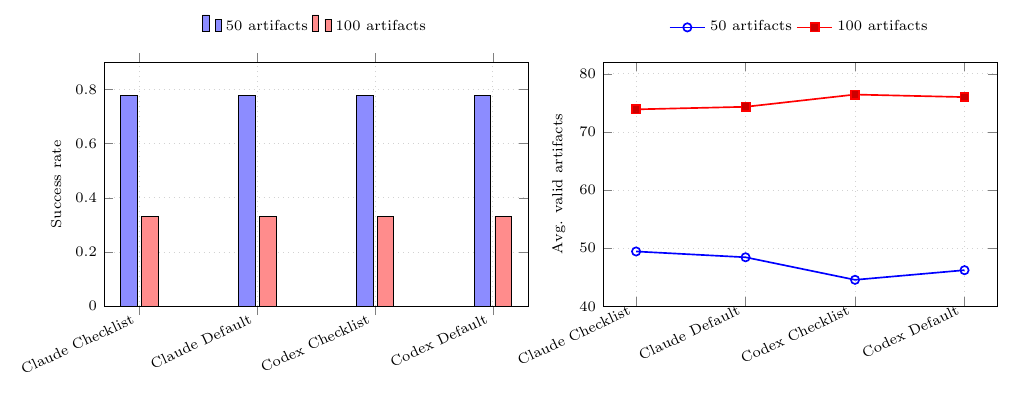}
\caption{Black-box frontier-agent \reposcan{} high-target evaluation. Each condition aggregates nine task
instances.
Both black-box harnesses solve many 50-artifact tasks, but success drops uniformly at 100 artifacts
even when average valid counts remain high.}
\label{fig:commercial-reposcan-summary}
\end{figure*}

\begin{center}
\begin{minipage}{\columnwidth}
\centering
\scriptsize
\setlength{\tabcolsep}{2pt}
\begin{tabular}{lrrrrr}
\toprule
Agent & $N$ & Pairs & Succ. $\Delta$ & Avg-valid $\Delta$ & Dup. $\Delta$ \\
\midrule
Claude Code (Sonnet 4.6) & 50 & 9 & 0.000 & 1.00 & 2.67 \\
Claude Code (Sonnet 4.6) & 100 & 9 & 0.000 & -0.44 & 13.44 \\
Codex CLI (\texttt{gpt-5.4}) & 50 & 9 & 0.000 & -1.67 & 0.22 \\
Codex CLI (\texttt{gpt-5.4}) & 100 & 9 & 0.000 & 0.44 & -8.78 \\
\bottomrule
\end{tabular}
\captionof{table}{Checklist-minus-default paired deltas over matched task and agent. The checklist prompt
changes duplicate and tool-use behavior but does not improve success in this evaluation.}
\label{tab:commercial-checklist-delta}
\end{minipage}
\end{center}

Figure~\ref{fig:commercial-reposcan-summary} shows that frontier-agent performance improves the
absolute success rate without eliminating the high-count persistence problem. Both black-box agents solve
seven of nine $N=50$ tasks under both prompts,
showing that the search-and-submit interface and task construction are usable. At $N=100$, however, every
agent-prompt condition drops to three of nine successes. The checklist prompt has zero paired
success delta for both agents and target counts. Generic progress reminders alone therefore do not
guarantee robust high-target performance; the remaining gap points back to mechanisms that track
verifier-aligned progress explicitly.

\end{document}